\documentclass[runningheads]{llncs}
\usepackage[T1]{fontenc}
\usepackage{graphicx}
\usepackage{adjustbox}
\usepackage{booktabs}
\usepackage{multirow}
\usepackage{amsmath,amssymb,amsfonts}
\usepackage{tabularx}
\begin{document}
\title{Bridging Semantic Gaps in RAG through Generated Context Knowledge Fusion}
\titlerunning{Knowledge-Aware Semantic Bridging}
%

\author{Xinkai Du\inst{1,2} \and
Chao Lv \inst{1} \and
Yalin Sun \inst{3}  \and
Quanjie Han \inst{1} \and
Lei Yao \inst{1} \and
Maosong Sun \inst{2} \thanks{Corresponding Author} }
%
%
\institute{Beijing Wanlian Zhilian Technology Corporation Limited, Beijing, China\\
\email{\{duxinkai, lvchao, hanquanjie, yaolei\}@wanlianyida.com}
\and
Department of Computer Science and Technology, Tsinghua University, Beijing, China\\
\email{sms@tsinghua.edu.cn}\\ \and
Sunshine Digital Intelligence Tech Co., Ltd., Beijing, China\\
\email{sunyalin-ghq@sinosig.com}}

\maketitle              
\begin{abstract}
Retrieval-Augmented Generation has established itself as a fundamental framework in natural language processing, seamlessly integrating information retrieval with the generative capabilities of large language models. However, this process is fundamentally constrained by a critical challenge: semantic space mismatch between queries and retrieved contexts. We propose Knowledge-Aware Semantic Bridging (KASB), a novel framework that improves passage selection quality through semantic space alignment between queries and retrieved documents through intelligent knowledge fusion. Our approach leverages the complementary strengths of generative and retrieval-based knowledge through a multistage process that enhances both
relevance and accuracy. We evaluate KASB on three popular open-domain Question Answering datasets to demonstrate the effectiveness of our approach.

\keywords{Large Language Model  \and Direct Preference Optimization \and Retrieval Augmented Generation \and Question Answering .}
\end{abstract}
\section{Introduction}
Retrieval-Augmented Generation (RAG) \cite{lewis2020retrieval} has become a cornerstone framework in natural language processing, combining information retrieval with the generative power of large language models (LLMs). The standard RAG pipeline consists of three stages: retrieval, re-ranking, and generation. In the retrieval phase, relevant documents are identified based on semantic similarity between the query and a knowledge corpus. However, this process is hindered by a key limitation: semantic space mismatch between queries and retrieved contexts. This mismatch arises from differences in linguistic
expression, abstraction levels, and terminology, even for semantically aligned content \cite{izacard2021leveraging}.

LLMs having acquired vast world knowledge through self-supervised pretraining \cite{brown2020language}, can generate contextually coherent responses during question answering. Yet, this generative capability is prone to hallucination: producing plausible but factually inaccurate information \cite{ji2023survey}. In contrast, retrieved knowledge from curated corpora tends to be more factually reliable but often lacks fine-grained relevance to the query. This trade-off between the relevance of generated content and the accuracy of retrieved content remains a fundamental challenge in current RAG systems.

We propose Knowledge-Aware Semantic Bridging (KASB), a novel framework that dynamically aligns the semantic spaces between queries and retrieved documents via adaptive knowledge fusion. Our method exploits the complementary advantages of generative knowledge and retrieval-based knowledge in a multi-stage paradigm, to substantially boost semantic relevance and factual accuracy simultaneously.

Our contributions can be summarized as follows:

\begin{itemize}
    \item We propose KASB, a reinforcement learning-enhanced methodology for dynamically aligning query and document representations. 
   \item We introduce an unsupervised LLM generated context based evidence selector within the RAG framework, which is an effective approach that leverages explicit hypothetical document as signals
to align the preferences of different components.

\item Extensive experiments are conducted on three
open domain QA datasets to demonstrate the effectiveness of KASB.
\end{itemize}

\begin{figure*}[h]
\centerline{\includegraphics[width=0.9\textwidth]{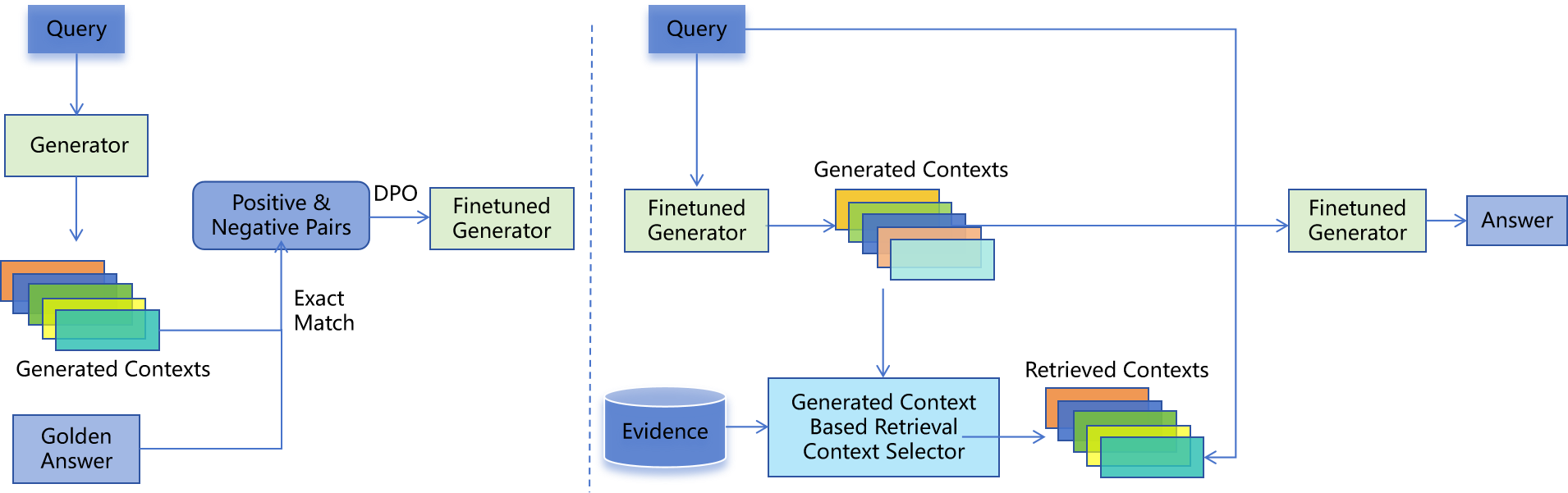}}
\caption{Overview of the KASB Framework.  A finetuned generator is derived via the DPO algorithm on the training set to produce a more effective generative context generator, which then serves in the test phase. During inference, the finetuned generator first generates query-aligned generative contexts $\mathcal{G}$; meanwhile, the evidence (representing the complete retrieved knowledge base) is processed through the generated context-based selector to identify relevant retrieved contexts $\mathcal{C}$, which are then fused with the generated contexts $[\mathcal{C};\mathcal{G}]$ for final answer generation. } 

\label{BRMGR}
\end{figure*}

\section{Related Work}
\label{sec:prior}
In knowledge-intensive tasks, the retrieve-then-read paradigm has been dominant \cite{lewis2020retrieval,karpukhin2020dense,izacard2021leveraging}. Early approaches relied on sparse retrievers such as BM25, while recent advances leverage dense contextualized representations, exemplified DPR \cite{karpukhin2020dense}, which significantly outperform traditional methods. Various strategies have since been explored to integrate the retrieved passages with large language models  to improve the generation of answers \cite{liu2023exploring}.

LLMs have also been employed to generate auxiliary knowledge for downstream reasoning \cite{liu2022generated}. Building on this capability, recent work has explored using LLM-generated knowledge to enhance retrieval, particularly when relevant information is sparse or insufficiently covered by a fixed corpus. This has led to the emergence of the generate-then-read framework, which generates query-specific contexts instead of retrieving them from a fixed corpus \cite{yu2022generate}. Methods like HyDE \cite{gao2023precise} further refine retrieval by generating hypothetical document embeddings for dense indexing.

Building on retrieval-augmented and generation-augmented paradigms, recent work has explored hybrid approaches to mitigate their respective limitations: retrieved passages may be irrelevant, while generated ones may be plausible but inaccurate. To address this, \cite{zhang2023merging} proposes merging passages based on compatibility scoring, and \cite{du2025improving} introduces an unsupervised bi-reranking framework for fusing generated and retrieved knowledge. Our method integrates the complementary strengths of both paradigms through a multistage process that improves relevance and accuracy through a lightweight embedding-space selection mechanism.

KASB differs from HyDE \cite{gao2023precise} in that HyDE primarily generates a hypothetical document to construct a retrieval representation, whereas KASB starts from an arbitrary retriever's candidate set and uses multiple generated contexts to select real evidence after retrieval. Compared with COMBO \cite{zhang2023merging} and BRMGR \cite{du2025improving}, KASB uses generated contexts not only for fusion but also as a query-conditioned signal for passage scoring and adaptive evidence selection.

\section{KASB: Knowledge-Aware Semantic Bridging Guided by Generated Context}
\label{sec:format}
\subsection{Generated Context Generation}
RAG inherently faces a semantic space mismatch between queries and retrieved contexts. This gap arises from discrepancies in linguistic expression, abstraction levels, and terminology usage, even when the core semantics are aligned. LLMs having internalized extensive world knowledge through self-supervised pretraining, can generate hypothetical relevant contexts tailored to the query’s semantic style. These generated contexts act as a semantic bridge to connect the query with potentially misaligned retrieved documents, addressing the fundamental limitation of traditional RAG’s retrieval phase.

For each query $q$, an LLM is prompted to generate a context $\mathcal{G}$, formulated as
\begin{equation}
\mathcal{G} = LLM(q,\text{prompt})
\end{equation}

\subsection{Preference-Tuned Generated Context Generator}
While LLMs can generate contexts well-aligned with user queries, their outputs frequently suffer from hallucinations, plausible but factually incorrect content that significantly undermine the reliability of downstream retrieval and generation processes \cite{ji2023survey,maynez2020faithfulness}. Traditional approaches to mitigating this issue often employ Reinforcement Learning from Human Feedback (RLHF) \cite{ouyang2022training}, which requires training and maintaining a separate reward model, introducing additional complexity and computational overhead.

Direct Preference Optimization (DPO) \cite{rafailov2023direct} provides an efficient alternative by directly optimizing the policy from preference pairs without an explicit reward model. To enhance the contextual generation capability of our language model, we fine-tune the generator on the training split of each dataset using preference pairs constructed through an automated process. 

Specifically, we leverage the Exact Match (EM) metric \cite{rajpurkar2016squad} to evaluate generated contexts against ground-truth answers, designating contexts containing the correct answer as positive samples and those without as negative samples. This creates a training signal that explicitly encourages factually accurate context generation while maintaining semantic alignment with the query. The generator is subsequently optimized through DPO, resulting in more reliable and factually grounded contexts for semantic bridging without requiring human-annotated preference data.

\begin{equation}
L_{\text{DPO}}(\pi_\theta; \pi_{\text{ref}}) = -\mathbb{E}_{(q, y_w, y_l) \sim \mathcal{D}} [\log\sigma(\beta f(q, y_w, y_l))]
\end{equation}

\begin{equation}
f(q, y_w, y_l)=\log \frac{\pi_\theta(y_w|q)}{\pi_{\text{ref}}(y_w|q)} -\log \frac{\pi_\theta(y_l|q)}{\pi_{\text{ref}}(y_l|q)}
\end{equation}

where $\pi_{\theta}$ denotes the model to be optimized, $\pi_{\text{ref}}$ represents the original model to generate question contexts, and $\theta$ stands for the model optimization parameters.

This fine-tuning process aligns the LLM’s generated contexts with factual accuracy, ensuring they serve as reliable signals for subsequent retrieval.

\subsection{Unsupervised Generated Context Based Retrieved Context Selection}

Retrieved contexts from curated corpora are factually reliable but often contain irrelevant noise. Given the fine-tuned knowledge-generating model's efficacy in producing query-relevant context, we posit that strong alignment between retrieved documents and this generated context serves as a reliable indicator of answer adequacy in top-ranked results.

Formally, we implement the document encoder $enc_d$ as a dual-encoder architecture, where for each query $q_j$, the similarity between retrieved contexts and the fine-tuned model's generated knowledge is computed by:
\begin{equation}
    \mathcal{S}(g_i, e_j) = <enc_d(g_i),enc_d(e_j)>
\end{equation}

All the related context for query $q_j$ is retrieved as follows:

\begin{equation}
    E_v = \left\{ \arg \max_{e_j \in E} \mathcal{S}(g_i, e_j) \mid g_i \in \mathcal{G} \right\}
\end{equation}

where $\mathcal{G}$ denotes all the generated context information produced by fine-tuning the large model for the query $q_j$, and $E_v$ represents the retrieved context information corresponding to the query $q_j$.

\subsection{Generated Context and Retrieved Context Combination}
To determine the optimal number of retrieved documents, we introduce an adaptive top-k selection mechanism based on similarity score distribution analysis. For each query, we first compute the centroid of generated context embeddings:

\begin{equation}
\overline{g} = \frac{1}{|\mathcal{G}|} \sum_{g_i \in \mathcal{G}} \text{SBERT}(g_i)
\end{equation}

where $\mathcal{G}$ denotes the set of contexts generated by the fine-tuned model, and SBERT refers to the pre-trained sentence embedding model from the sentence-transformers library.

Subsequently, we calculate cosine similarities between the centroid $\overline{g}$  and all retrieved contexts, yielding an ordered sequence \(\{s_1, s_2, \ldots, s_n\}\) sorted in descending order. To identify the optimal cutoff point, we analyze the first-order differences of these similarity scores:
\begin{equation}
\Delta_k = s_k - s_{k + 1}
\end{equation}

We use z-score to standardize all the first-order difference values. We select the first index \(k^*\) where \(\Delta_k\) significantly deviates from the mean z-score. If no such decline is found, we will use the second-order difference method to find the point with the maximum curvature to obtain the corresponding index \(k^*\).

Finally, the selected top-\(k^*\) retrieved contexts $\mathcal{C}$ and the generated contexts produced by the fine-tuned large model are taken as a whole as the context information of the query.  The resulting context is then fed into the fine-tuned large language model to generate the answer.

\section{Experimental Setup}
\label{sec:pagestyle}
To demonstrate the effectiveness of our method, we evaluate it on three popular open-domain Question Answering datasets.

\textbf{Datasets:} Following prior work~\cite{du2025improving},  we conduct experiments on three widely-used open-domain Question Answering (QA) datasets: TriviaQA~\cite{joshi2017triviaqa}, Natural Questions (NQ)~\cite{kwiatkowski2019natural}, and WebQuestions (WebQ)~\cite{berant2013semantic}.

\textbf{Evaluation Metric:} For question answering, we employ the exact match (EM) score \cite{rajpurkar2016squad}. For the retrieval task, we adopt the standard top-K retrieval EM metric, which measures the proportion of questions where at least one passage among the top-K retrieved ones contains a text span that exactly matches the human-annotated answer.

\textbf{Baselines:} 
Comparable to \cite{zhang2023merging}, we employ single and two knowledge sources in comparison experiments to demonstrate our method's effectiveness.
\begin{itemize}
\item \textbf{Retri-Only}: Utilizes only the original retrieved knowledge directly as the input for the downstream reader model.
\item \textbf{Gen-Only}: Adopts only the original LLM-generated knowledge as the sole input for the reader model.
\end{itemize}
\begin{itemize}
\item \textbf{HyDE}: A classic retrieval-augmented baseline that generates a hypothetical document for each query, encodes it into a dense representation, and uses this representation to retrieve relevant real documents \cite{gao2023precise}.
\item \textbf{COMBO}: Matches LLM-generated passages with their retrieved counterparts to form compatible knowledge pairs and fuses the paired knowledge for open question answering \cite{zhang2023merging}.
\item \textbf{BRMGR}: An unsupervised Bi-Reranking for Merging Generated and Retrieved knowledge method, which optimizes the fusion of two knowledge sources via bi-directional reranking of retrieved and generated passages \cite{du2025improving}.
\end{itemize}

\textbf{DPO Training Details:} We implement DPO optimization using the LLaMA Factory framework \cite{zheng2024llamafactory} with a learning rate of $5 \times 10^{-6}$, batch size 4, $\beta=0.1$, and 3 training epochs. The reference model is the original Meta-Llama-3.1-8B-Instruct. Positive and negative contexts are constructed automatically from answer containment: a generated context containing the ground-truth answer is preferred over one that does not. We therefore refer to these as \emph{automatically constructed answer-containment preferences}, rather than human preferences.

We use Meta-Llama-3.1-8B-Instruct with the prompt ``Question: \{question\} Provide effective generated contexts for retrieving relevant information.'' Five relevant contexts are generated per query and encoded with \texttt{ms-marco-MiniLM-L6-v2}.

For each query, if the generated contexts include at least one positive and one negative instance, the query and the corresponding positive/negative contexts are formatted into ShareGPT style. This construction is scalable but can produce false positives for polysemous strings and false negatives for aliases or paraphrases. The data statistics for DPO training are summarized in Table \ref{data_statistics}:

 \begin{table}[htb]
\centering  
\caption{Datasets statistics.}\label{datasets_statistics}
\setlength{\tabcolsep}{10pt}
\renewcommand{\arraystretch}{1.2}
\begin{tabular}{lccc}
\hline
\rule{0pt}{15pt}
&Datasets & Train & Dev \\
\hline
&TriviaQA & 75675 & 8750 \\
&NQ & 91334 & 10039 \\
&WebQ & 3906 & - \\
\hline
\end{tabular}
\label{data_statistics}
\end{table}

\section{Experimental Results}
\label{sec:typestyle}

\subsection{Main Results}
\label{sec:majhead}
\subsubsection{Finetuned LLM Generation Ability}
\begin{figure}[htb]
  \centering
  \centerline{\includegraphics[width=\textwidth]{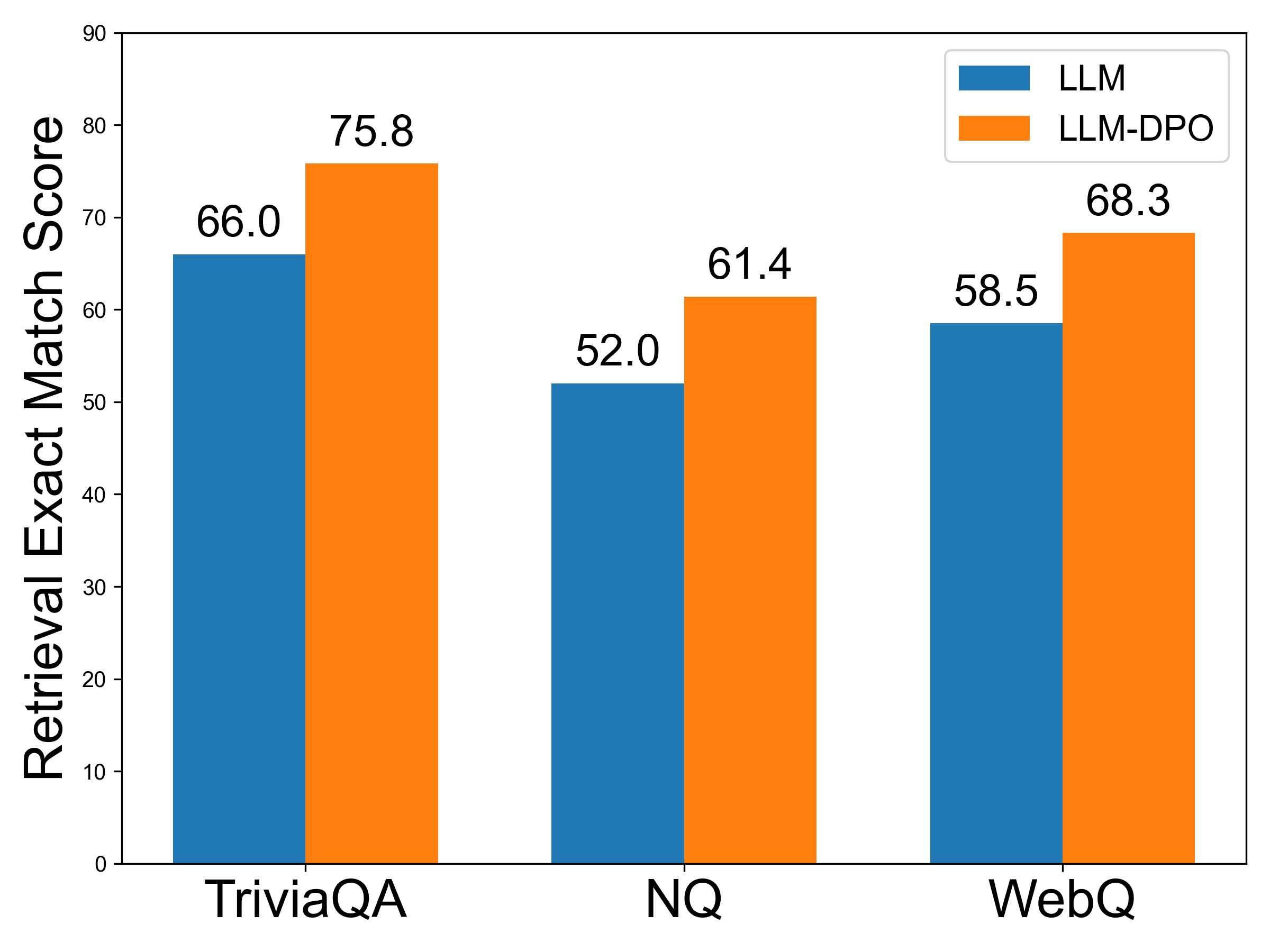}}
\caption{ Average retrieval exact match score obtained by finetuned vs original LLM generator on generated contexts.}
\label{finetuned_llm}
\end{figure}

To enhance the LLM's capability in generating accurate and relevant contextual knowledge, we fine-tuned the generator using the DPO algorithm with carefully constructed positive and negative sample pairs. As demonstrated in Figure \ref{finetuned_llm}, the finetuned LLM generator achieved substantial improvements in average retrieval exact match scores, with absolute gains of 14.85\%, 18.08\%, and 16.75\% on the TriviaQA, NQ, and WebQ test sets, respectively.

\subsubsection{Question Answering }
In order to assess the performance of our method in open question answering, we employ the finetuned LLM generator as the reader model. The results of this evaluation are presented in Table \ref{oqa_em_result}.

Table  \ref{oqa_em_result} clearly illustrates that our proposed method KASB achieves the strongest overall performance across all three datasets.
Specifically, methods leveraging two knowledge sources (retrieved and generated) consistently outperform those relying on a single knowledge source alone. For example, even the weakest two-source method BRMGR surpasses the best single-source method on NQ and WebQ, highlighting the advantage of fusing complementary retrieved and generated knowledge for open question answering.

Among the two-source baselines, KASB delivers the most substantial gains. This underscores the effectiveness of our knowledge alignment and semantic boosting strategies in optimizing the fusion of dual knowledge sources.
Additionally, when examining single-source methods, Gen-Only outperforms Retri-Only on TriviaQA and NQ, while Retri-Only exhibits stronger performance on WebQ, revealing the complementary strengths of generated and retrieved knowledge across datasets with different characteristics.

\begin{table}
\begin{center}
{\caption{Exact match scores on test dataset.}\label{oqa_em_result}}
\begin{tabular}{p{2.8cm}p{2cm}p{2cm}p{2cm}}
\hline
\rule{0pt}{12pt}
Methods&\textbf{TriviaQA}&\textbf{NQ}&\textbf{WebQ}\\
 \hline
\textit{\scriptsize Single Knowledge}&\multicolumn{3}{c}{}\\
Retri-Only&62.2&48.6&48.3\\
Gen-Only&67.5&50.3&41.6\\
\textit{\scriptsize Two Knowledge}&\multicolumn{3}{c}{}\\
HyDE &72.3&53.4&54.6\\
COMBO &74.6&54.2&53.0\\
BRMGR & 68.6 & 52.2 & 53.4 \\
KASB & \textbf{75.4} & \textbf{59.6} & \textbf{61.2} \\
 \hline
\end{tabular}
\end{center}
\end{table}

\subsection{Ablation Analysis}
\label{sec:print}
When both retrieved and generated knowledge are available, we use them as contextual input to prompt the fine-tuned LLM, which then directly generates the answer. The model’s accuracy is evaluated using exact match (EM) score against the ground truth.

\subsubsection{Retrieval Performance Boost via Semantic Bridging}
To assess the improvements in retrieval performance brought about by semantic bridging, we conduct dedicated experiments where passages are retrieved using both unsupervised and supervised retrievers. The unsupervised retrievers selected for our experiments include BM25, MSS \cite{sachan2021end}, and Contriever \cite{izacardunsupervised}, whereas the supervised retrievers consist of DPR \cite{karpukhin2020dense} and MSS-DPR \cite{sachan2021end}.

Retrieval accuracy at Top-20 and Top-100 on the test sets of the three datasets, evaluated on the basis of the top-1000 retrieved passages, is reported in Table \ref{tab:retriever_performance}. As illustrated in the table, our proposed KASB method yields consistent improvements in retrieval performance across all retrievers, with notably substantial gains achieved for unsupervised models including BM25, MSS \cite{sachan2021end}, and Contriever \cite{izacardunsupervised}.

\begin{table*}[ht]
    \centering
    \caption{Top-{20, 100} retrieval accuracy on the test set of datasets for the top-1000 retrieved passages.}
    \label{tab:retriever_performance}
    \begin{adjustbox}{}
    \begin{tabular}{l|cc|cc|cc}
    \toprule
    \textbf{Retriever} & \multicolumn{2}{c}{\textbf{TriviaQA}} & \multicolumn{2}{c}{\textbf{NQ}} & \multicolumn{2}{c}{\textbf{WebQ}} \\
     & Top-20 & Top-100 & Top-20 & Top-100 & Top-20 & Top-100 \\
    \midrule
    \multicolumn{7}{c}{\textit{Unsupervised Retrievers}} \\
    \midrule
    MSS & 67.2 & 79.1 & 60.0 & 75.6 & 49.2 & 68.4  \\
    MSS + KASB  & 81.3 & 85.0 & 77.3 & 81.5 & 62.8 & 75.8  \\
    \midrule
    BM25  & 76.4 & 83.2 & 62.9 & 78.3 & 62.4 & 75.5  \\
    BM25 + KASB  & 84.3 & 87.2 & 75.6 &84.3  & 65.6 & 77.4  \\
    \midrule
    Contriever  & 73.9 & 82.9 & 67.9 & 80.6 & 65.7 & 80.1  \\
    Contriever + KASB &85.7  &87.4  &81.2  &86.5 &70.1 &81.3  \\
    \midrule
    \multicolumn{7}{c}{\textit{Supervised Retrievers}} \\
    \midrule
    DPR  & 79.8 & 85.1 & 79.2 & 85.7 & 74.6 & 81.6  \\
    DPR + KASB  & 84.8 & 87.4 &83.5  &87.4  &78.2  & 84.2  \\
    \midrule
    MSS-DPR & 81.9 & 86.6 & 81.4 & 88.1 & 76.9 & 84.6  \\
    MSS-DPR + KASB &87.5  &88.3  & 84.3 &89.7  &81.4  & 85.3 \\
    \bottomrule
    \end{tabular}
    \end{adjustbox}
    
\end{table*}

\subsubsection{Effect of the finetuned Generated Context Generator}
Comparing KASB with the w/o DPO variant in Table \ref{tab:qa_em_result}, we observe that the fine-tuned LLM  achieves significantly higher performance across all three dataset compared to the baseline model without direct preference optimization. This indicates that the fine-tuning process, particularly through direct preference optimization enhances the model’s ability to generate high-quality responses. The substantial improvement in exact match scores suggests that the fine-tuned LLM aligns better with the automatically constructed answer-containment preferences and produces more accurate answers when integrated with retrieved and generated knowledge.
\begin{table}[t]
\centering
\caption{Question answering performance (Exact Match). GK and RK denote generated knowledge and retrieved knowledge, respectively. The combined knowledge results are computed by union of single knowledge sources.}
\label{tab:qa_em_result}
\begin{tabular}{p{2.8cm}p{2cm}p{2cm}p{2cm}}
\toprule
\textbf{Model} & \textbf{TriviaQA} & \textbf{NQ} & \textbf{WebQ} \\
\midrule
KASB & \textbf{75.4} & \textbf{59.6} & \textbf{61.2} \\
w/o GK & 69.9 & 54.9 & 56.3 \\
w/o RK & 73.2 & 56.8 & 59.5 \\
w/o DPO & 74.6 & 57.7 & 59.8 \\
\bottomrule
\end{tabular}
\end{table}

\subsubsection{Importance of the Retrieved and Generated Knowledge Fusion}

Comparing KASB with the ablated variants in Table \ref{tab:qa_em_result}, both knowledge sources contribute to performance. Removing generated knowledge (GK) causes a larger drop than removing retrieved knowledge (RK), while removing DPO has a smaller effect. This supports the interpretation that the main benefit comes from the interaction of generated-context guidance, evidence selection, and two-source fusion rather than from DPO alone.

\subsection{Parameter Sensitivity Analysis}
\label{sec:sensitivity}

\begin{figure}[ht]
\centering
\includegraphics[width=\columnwidth]{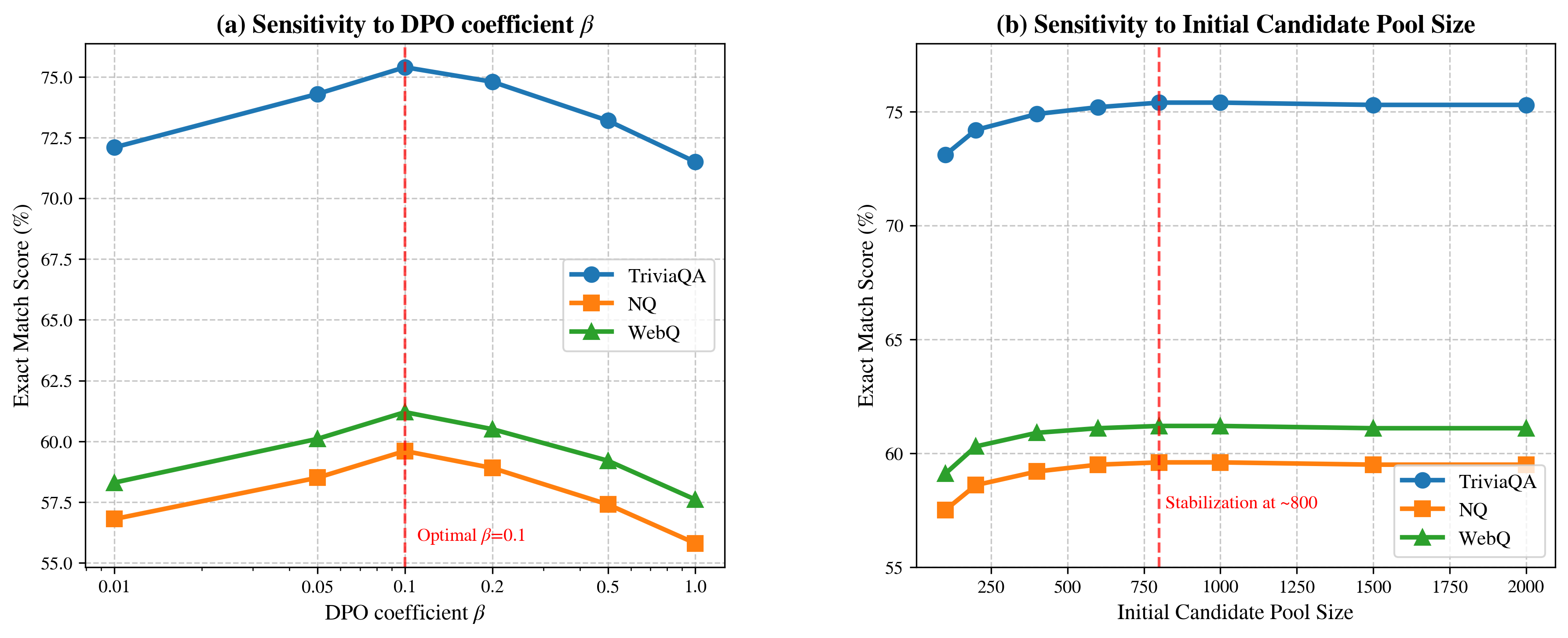}
\caption{Parameter sensitivity analysis showing the effect of DPO coefficient $\beta$ and initial candidate pool size on EM performance across datasets.}
\label{fig:sensitivity}
\end{figure}

Figure \ref{fig:sensitivity} examines the sensitivity of KASB to two critical hyperparameters: the DPO coefficient $\beta$
and the initial number of retrieved candidates. Performance peaks at 
$\beta=0.1$ suggesting that moderate preference strength optimally balances alignment with retained knowledge diversity. For retrieved candidates, performance stabilizes after 800 documents, indicating that KASB can work effectively with standard retrieval pool sizes without requiring exhaustive retrieval.

The adaptive top-$k$ mechanism yields average $k^{\star}$ values of 24.3 (TriviaQA), 18.7 (NQ), and 15.2 (WebQ). To directly examine whether adaptive selection is preferable to a globally fixed cutoff, we further compare fixed $k\in\{5,10,20,50\}$ in Table~\ref{tab:fixed_k}. The fixed-$k$ results show a consistent non-monotonic trend: increasing $k$ initially improves performance by adding useful evidence, while overly large $k$ introduces more less-relevant passages and degrades performance. In contrast, the adaptive strategy achieves the best result on all three datasets. This indicates that a single global $k$ is suboptimal because the appropriate amount of evidence varies with the query-specific score distribution.

\begin{table}[ht]
\centering
\caption{Fixed-$k$ versus adaptive evidence selection. EM is reported on the three QA datasets; Avg. is the mean across datasets.}
\label{tab:fixed_k}
\begin{tabular}{p{3cm}p{2cm}p{2cm}p{2cm}p{2cm}}
\toprule
\textbf{Selection} & \textbf{TriviaQA} & \textbf{NQ} & \textbf{WebQ} & \textbf{Avg.} \\
\midrule
Fixed-$k$, $k=5$  & 73.2 & 58.2 & 59.2 & 63.5 \\
Fixed-$k$, $k=10$ & 73.8 & 58.5 & 59.8 & 64.0 \\
Fixed-$k$, $k=20$ & 74.2 & 58.9 & 59.7 & 64.3 \\
Fixed-$k$, $k=50$ & 73.5 & 58.1 & 59.4 & 63.7 \\
Adaptive $k^\star$ & \textbf{75.4} & \textbf{59.6} & \textbf{61.2} & \textbf{65.4} \\
\bottomrule
\end{tabular}
\end{table}

\subsection{Discussion}
\label{sec:limitation}

Despite KASB's effectiveness, several limitations warrant discussion. First, the quality of semantic bridging is constrained by the generative model's knowledge coverage and accuracy; generated contexts may be outdated, redundant, hallucinated, or conflicting with retrieved evidence. Second, our DPO preferences are automatically constructed from answer containment rather than human judgments, so they may reward surface overlap and do not establish semantic correctness; answer-masked or answer-absent evaluation would be a useful diagnostic. Third, the fixed-$k$ comparison shows that performance first improves as more evidence is included and then declines when overly large cutoffs introduce less-relevant passages, supporting the use of query-specific adaptive selection. Fourth, we do not explicitly perform deduplication or contradiction resolution between generated and retrieved knowledge. Finally, the current evaluation is limited to single-turn open-domain QA, and we do not claim cross-dataset generalization of the DPO-tuned generator without additional transfer experiments.

\section{Conclusion}
\label{sec:conclusion}

In this paper, we propose Knowledge-Aware Semantic Bridging (KASB), a lightweight post-retrieval framework that uses generated contexts to guide evidence selection and fuse selected retrieved passages with generated knowledge. Experiments on three single-turn open-domain QA datasets show consistent improvements in retrieval accuracy and answer correctness across diverse retrievers. The results support generated-context-guided evidence selection as a practical system-level integration for improving RAG.

\subsection*{Acknowledgements}
This work is supported by Beijing Municipal Science and Technology Plan Project(Z241100001324025).

%
%
%
\bibliographystyle{splncs04}
\bibliography{strings}
%




\end{document}